\pdfoutput=1
\documentclass[11pt]{article}

\usepackage[preprint]{acl}

\usepackage{times}
\usepackage{latexsym}
\usepackage{booktabs}
\usepackage{tabularx}

\usepackage[T1]{fontenc}
\usepackage[utf8]{inputenc}

\usepackage{microtype}

\usepackage{inconsolata}

\usepackage{graphicx}

\usepackage{ascmac}  % to use itembox
\usepackage{amsmath}  % to use equation
\usepackage{cleveref}  % to use Cref
\usepackage{multirow}  % to use multirow
\usepackage{booktabs}  % to use rules

\title{Augmenting and Refining NER Datasets through LLMs}
\author{Yuji Naraki \thanks{Equal contributions. \\Correspondence to: \\Yuji Naraki <yuji.1277@gmail.com>, \\Ryosuke Yamaki <yamaki.ryosuke@em.ci.ritsumei.ac.jp>, \\Hiroki Naganuma <hiroki.naganuma@proplace.jp>.} \\
Independent Researcher
\vspace{-20mm}  % 個別の調整
\And
Ryosuke Yamaki$^*$ \\
Ritsumeikan University / ProPlace Inc
\vspace{-20mm}  % 個別の調整
\AND
Yoshikazu Ikeda \\
Osaka University / ProPlace Inc \\
\vspace{-20mm}  % 個別の調整
\And
Takafumi Horie \\
Ritsumeikan University
\vspace{-20mm}  % 個別の調整
\And
Kotaro Yoshida \\
Science Tokyo
\vspace{-20mm}  % 個別の調整
\AND
Ryotaro Shimizu \\
ZOZO Research
\vspace{-20mm}  % 個別の調整
\And
Hiroki Naganuma \\
Mila, Université de Montréal / ProPlace Inc
\vspace{-20mm}  % 個別の調整
}
\begin{document}
\vspace{-5mm}  % タイトルと著者情報の間の余白を調整
\maketitle
\vspace{-5mm}  % 著者情報と本文の間の余白を調整
\begin{abstract}
Traditional named entity recognition (NER) dataset annotation methods often suffer from high costs and inconsistent quality. This study proposes a hybrid annotation approach that combines human effort with large language models (LLMs) to reduce noise due to the missing labels and improve NER model performance cost-effectively. A label-mixing strategy is also introduced to address class imbalance. Our experiments demonstrate that the proposed method outperforms traditional approaches, even under limited budget constraints.
\end{abstract}

\section{Introduction}
% In the field of Natural Language Processing (NLP), Named Entity Recognition (NER) is a critical task that involves identifying and classifying named entities in a given text into predefined categories, including persons, organizations, locations, temporal expressions, quantities, monetary values, and percentages~\citep{mohit2014named}. 
% NER is integral to a wide range of applications, from basic information retrieval and content classification to more complex tasks such as question answering and machine translation~\citep{10.1145/1571941.1571989, Aliod2006NamedER, Babych2003ImprovingMT}. 
% The enhancement of NER model performance not only directly contributes to the improvement of broad NLP applications, but also plays a crucial role in enhancing our understanding and engagement with text-based data~\citep{Cheng2019AttendingTE, Etzioni2005UnsupervisedNE}.

Named entity recognition (NER) is a critical task that involves identifying and classifying named entities in a given text into predefined categories \citep{mohit2014named}.
% NER methods leveraging the in-context learning capabilities of Large Language Models (LLMs) have been proposed, demonstrating high performance close to that of models specialized in NER by carefully selecting the examples provided to LLMs~\citep{Wang2023GPTNERNE}.
% %This indicates the effectiveness of utilizing LLMs' advanced text comprehension abilities for NER tasks. 
% %On the other hand, inference with LLMs requires significant computational resources and involves lengthy latency before obtaining inference results, making them unsuitable for applications requiring the rapid processing of huge amounts of text data. 
% However, inference using LLMs is costly, which may limit their practicality for real-world applications that demand the rapid processing of large-scale text data.
% %Therefore, we argue that it is crucial to develop approaches that leverage LLMs' language understanding capabilities while requiring fewer computational resources and achieving faster NER task processing. 
% This study focuses on improving the quality of datasets used for training NER models with the aid of LLMs, aiming to enhance the performance of smaller, faster, dedicated NER models.
NER models' performance relies heavily on the quality of annotations in the datasets~\citep{Mitkov2000TheIO}. 
Currently, most datasets employed for training NER models are annotated by humans.
This manual annotation process is fraught with several challenges, such as annotation errors, time and financial costs~\citep{Katiyar2018NestedNE,fort2011amazon}.

\begin{figure}[t]
    \centering
    \includegraphics[width=1.0\linewidth]{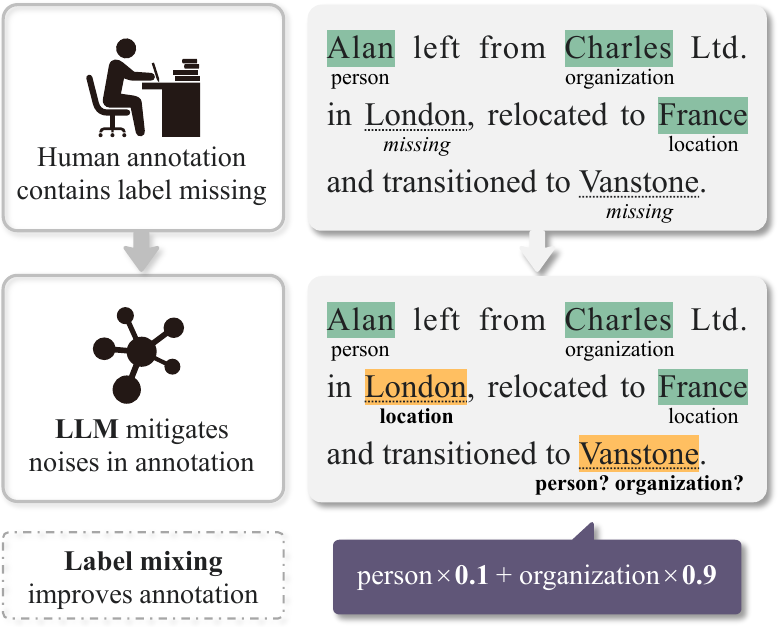}
    \caption{LLM-based annotation and label mixing.}
    \vspace{-3.5ex}
    \label{fig:method}
\end{figure}

This study introduces a novel method to address the challenges of manual annotation by leveraging large language models (LLMs) as annotators. The approach aims to improve dataset quality, which enhances the performance of NER models.
% In comparison to human annotators, LLMs are considered as consistent and efficient in the annotation process. 
% While human annotators are considered as subject and fatigue, LLMs are considered as consistent and efficient in the annotation process. 
% We believe employing LLMs as annotators has some benefits.
Using LLMs for annotation offers several advantages.
% Human annotators are prone to subjectivity and fatigue, which can affect the quality of their work.
Human annotators can be subjective and suffer from fatigue, which may reduce accuracy.
In contrast, LLMs offer consistency and efficiency throughout the annotation process.
Furthermore, while employing LLMs as predictors leads to increasing costs with repeated predictions, employing them as annotators only requires a one-time cost during the annotation process.
% Afterward, small models, such as BERT, can be used as the predictor, providing adequate performance for the NER task.
Afterward, training a predictor based on small models, such as BERT, with the improved dataset can provide adequate performance for the NER task.

% This study proposes a novel approach to address these challenges associated with manual annotation, intending to enhance the quality of datasets and thereby strengthen the performance of NER models.
% This study proposes a novel approach to address these challenges associated with manual annotation by utilizing LLMs as annotators and validate the enhancement of the quality of datasets and thereby strengthen the performance of NER models.
% Compared to human, LLM has the higher consistency and efficiency in terms of annotation. 
% LLMをpredictorとして使うと推論ごとにコストが嵩むが、annotatorとして使うなら、アノテーション時に一度行えば十分であり、推論は比較的小さなBERTなどのモデルを使えばNERタスクは十分である。
% Using LLM as predictor, the cost of each inference is high, but if it is used as annotator, it is sufficient to do it once during annotation, and inference is sufficient for the NER task if a relatively small model such as BERT is used.
% Our approach leverages LLMs to supplement the missed annotations in manually annotated datasets,
% %that have been manually annotated (where the presence of missed annotations is assumed)
% as shown in \Cref{fig:method}.
%By utilizing a hybrid approach that combines manual annotations with annotations generated by large LLMs, we aim to recover from the noise introduced by missed annotations in the dataset, thereby enhancing the quality of the datasets used for training NER models at a low cost and in an automated manner.
Our approach using LLMs as annotators reduces the noise of missed labels by integrating manual and LLM-based annotations (as shown in \Cref{fig:method}), thus improving the quality of the data set for NER model training in a cost-effective and automatic manner.
We find that LLMs tended to assign multiple named entity labels to a single expression.
Additionally, an imbalance in data volume across labels leads to performance disparities among labels. 
To address this, we introduce a label mixing technique that integrates multiple named entity labels, reducing the variance of performance among labels and enhancing the model’s robustness.

% In our proposed method, a detailed analysis of the annotations by LLMs revealed that a single expression could be assigned multiple named entity labels.
%In addition, there is an imbalance in the data volume per label, which in turn causes performance disparities among labels in the NER model. 
% This is particularly critical in the ``Miscellaneous'' category, where such inequalities are more pronounced.
% To address these issues, we implemented a label mixing technique for the multiple assigned named entity labels, thereby ensuring the robustness of the NER model by mixing these labels.

%In the experiments, the CoNLL03~\citep{tjong-kim-sang-de-meulder-2003-introduction} and WikiGold~\citep{balasuriya-etal-2009-named} datasets were enhanced using our proposed hybrid annotation method combining manual human efforts and LLMs, and the improved datasets were then used to train a BERT-based~\citep{devlin-etal-2019-bert} NER model.
% Comprehensive evaluation experiments validated our approach's performance, demonstrating
% %The experimental results demonstrated that our proposed approach could
% the high performance in training NER models with datasets containing noise, such as missed annotations, despite requiring minimal financial cost.
Comprehensive evaluation experiments validated the effectiveness of our approach, demonstrating its ability to train NER models on noisy datasets, such as those with missed labels, while requiring minimal financial cost.

% The contributions of this study can be summarized in three main points:
% \begin{itemize}
%     \setlength{\itemsep}{0pt} % アイテム間のスペース
%     \setlength{\parsep}{0pt}  % 段落間のスペース
%     \item We propose a hybrid annotation method combining manual human efforts and LLMs for NER datasets, automatically recovering from noise such as missed annotations contained within the datasets (\Cref{fig:hybrid}).
%     \item By incurring only minimal financial cost, we enhance the performance of NER-specialized models (\Cref{tab:budget_comparison}).
%     \item In response to the issue of imbalance among named entity labels arising from LLM-based annotations, we propose a label mixing technique, ensuring the robustness of the NER model (\Cref{fig:label_mixing}).
% \end{itemize}

% \input{latex/02_preliminary}
\section{Related Work}
In machine learning, the accuracy and reliability of the dataset used for model training are critical factors directly influencing the performance of the trained model.
In this context, research has been conducted to scrutinize the reliability of datasets used for training NER models. 
Research, such as the one presented in \citet{Han2023IsIE}, highlights the inherent errors in these datasets.
These inaccuracies range from mislabeled entities to inconsistent categorization, impacting the overall performance and reliability of NER models \citep{mishra2020assessing, Mitkov2000TheIO}. 

% \citet{wang2023gptner} attempted to utilize LLMs for NER tasks as predictors instead of annotators and validated its effectiveness.
% They achieved the performance improvement by estimating the token range of entities to be sandwiched between special tokens (\textit{``@@, \#\#''}).

% 以下はintroに入れるべきこと。
% LLMs' advantages are high consistency and efficiency. 
% Conventional datasets with manual annotation tend to vary due to subjectivity and fatigue. 
% In addition, the cost of LLMs is much cheaper than that of human annotators.
% The disadvantage of their work is that the computational cost of inference is very high each time.
% While their approach directly solves the NER task, which differs, our approach includes augmenting and creating the NER dataset from scratch. 

% \subsection{Challenges in Automatic Annotation}

There are several challenges in automatic annotation. 
\citet{ghosh2023aclm} address the data imbalance problem through an attention map-based data expansion.
% ACLM (Attention-map aware keyword selection for Conditional Language Model fine-tuning), an attention map-based data expansion method to address the problem of data imbalance. 
% It should be noted, however, that this method does not take advantage of LLMs.
\citet{shen2023promptner} utilize multiple prompt templates to address the problem of multiple label assignments to the same entity.
\citet{shen2023diffusionner} and \citet{zhang2022bias} address the issues of noise and generative bias by utilizing causality-based methods. 
To address these problems, we propose a hybrid approach that combines manual and automatic LLM annotations. 
% % Specifically, we achieve the following three points:
% % 1) probabilistic integration of human and LLMs annotations to mitigate data class imbalance; 2) automatic elimination of label overlap using a label-mixing method; and 3) mitigation of noise/bias effects by combining human and LLMs.

% Other related studies in NER that must be mentioned are \cite{li2020survey} for a comprehensive review of NER deep learning methods, \cite{liu2021noisy} addresses noise from external knowledge bases, and \cite{wang2022miner} propose MINER (Mutual Information based Named Entity Recognition) for out-of-vocabulary support.

% Although the above diverse approaches address the challenge of NER, this study proposes a unique methodology that focuses on improving the reliability of the dataset itself and leads to better accuracy for NER tasks.

\section{Methodology}
\subsection{LLM-based Annotation}
% In linguistic data automation using LLM, the GPT-NER methodology, as delineated in \citet{wang2023gptner}, emerges as a paradigmatic example. This method profoundly influences our study, harnessing LLMs to facilitate automated annotation. Central to the GPT-NER strategy is the LLM's capability to discern and earmark specific entities within the text, encapsulating these segments with distinct character strings, as in \textit{``I live in @@Tokyo\#\#''}.

% [長沼書き直しは以下]
In the context of automating linguistic data using LLMs, the GPT-NER method, as described by \citet{wang2023gptner}, provides a key approach. This method plays a significant role in our study by leveraging LLMs for automated annotation. A core aspect of GPT-NER is the ability of the LLM to identify and mark specific entities within text, surrounding these with designated character strings, such as \textit{``I live in @@Tokyo\#\#''}.

The selection of few-shot examples is based on the approach outlined by \citet{wang2023gptner}, where examples are chosen from validation data using cosine similarity between token embeddings in the input text.

% For illustrative purposes, given an input such as \textit{``I live in Tokyo''}, wherein the objective is to pinpoint \texttt{LOC} (location) entities, the LLM's output would be \textit{``I live in @@Tokyo\#\#''}. This delineation effectively signifies the accurate detection of \textit{``Tokyo''} as a location entity, with the ``@@'' and ``\#\#'' serving as the entity's boundaries.
% Additionally, this approach empowers the LLM to apprehend both the intrinsic properties of entities and their interplay with the input-output schema, achieved through the integration of few-shot examples. In alignment with this methodology, we have formulated the prompt structure, as outlined in the box above.

% Within this structured prompt design, the [\textit{target entity type}] slot accommodates various entity types, including but not limited to \texttt{PER}, \texttt{ORG}, \texttt{LOC}, and \texttt{MISC}. Concurrently, the [\textit{definition target entity type}] slot is allocated for the explicit definition of these enumerated entity types. Moreover, the [\textit{Example $K$ Input}] and [\textit{Example $K$ Output}] slots are respectively reserved for the input text and its corresponding output text, collectively functioning as few-shot examples.

% [長沼書き直しは以下]
Additionally, this approach enables the LLM to understand both the inherent properties of entities and their relationships within the input-output framework by incorporating few-shot examples. Based on this methodology, we designed the prompt structure as described above.

In this prompt structure, the [\textit{target entity type}] slot is used for different entity types, such as \texttt{PER}, \texttt{ORG}, \texttt{LOC}, and \texttt{MISC}. The [\textit{definition target entity type}] slot is allocated for explicitly defining these entity types. The [\textit{Example $K$ Input}] and [\textit{Example $K$ Output}] slots are designated for the input text and its corresponding output, serving as few-shot examples.

\begin{figure}[t]
  \begin{itembox}{Instructions for the Large Language Model}
    \label{instruction_box}
    \small
    You are an excellent linguist.\par
    Identify [\textit{target entity type}] entities in given text.\par
    If the text contains no [\textit{target entity type}] entities, replicate the input text without any changes.\par
    Strictly adhere to the definition provided below.\par
    Definition of [\textit{target entity type}]:\par
    [\textit{definition target entity type}]\par
    \par
    [\textit{Example $1$ Input}]\par
    [\textit{Example $1$ Output}]\par
    [\textit{Example $2$ Input}]\par
    [\textit{Example $2$ Output}]\par
    $\cdots$\par
    [\textit{Example $K$ Input}]\par
    [\textit{Example $K$ Output}]
  \end{itembox}
  % \vspace{-6ex}
\end{figure}

% As this prompt dictates, the entity recognition process is methodically applied to each target entity type within the text targeted for annotation. In scenarios where this strategy involves multiple iterations of annotation for each type of entity, a single word may be labled with multiple different entity types. To resolve instances of such overlapping labels, the simplest method we adopt is to select one entity label from those multiple assignments randomly.

% The criterion for selecting few-shot examples is inspired by \citet{wang2023gptner}, entailing the selection of examples from validation data. This selection is predicated on the cosine similarity between token embeddings present in the input text.

% [長沼書き直しは以下]
The entity recognition process is systematically applied to each target entity type in the text for annotation. In cases where multiple iterations are required for different entity types, a single word may receive multiple labels. To handle such overlapping labels, we use a simple approach: one label is randomly selected from the multiple assignments.

The few-shot examples are chosen based on the method described by \citet{wang2023gptner}, which involves selecting examples from validation data based on the cosine similarity between token embeddings in the input text.

\subsection{Hybrid: LLM-Based and Manual Annotation}

% To enhance the quality of the NER dataset, we created Hybrid annotations by merging labels from LLMs with those generated manually. This merging process involved carefully analyzing the labels from both sources, ensuring consistency and accuracy. 
% The merged data combines all entity labels from the manual data and those from LLM-based annotation that do not overlap the same position as those from the manual data. Since the labels of the manual data are expected to be of higher quality than the ones from LLM-based annotation, they are merged in preference to the ones from LLM-based annotation.
% The merged dataset offers a more robust and diverse set of labels, mitigating biases and errors that might exist in either of the individual sources.

To enhance the quality of the NER dataset, we created hybrid annotations by merging labels from LLMs with manually generated ones. This process involved carefully analyzing both sources to ensure consistency and accuracy. Entity labels from the manual data were prioritized, with LLM-based labels added only when they did not overlap. As manual annotations are considered higher quality, they were preferred over LLM-generated ones. The resulting merged dataset offers a more robust and diverse set of labels, reducing biases and errors from either source.

\begin{figure}[t]
    \centering
    \includegraphics[width=1.0\linewidth]{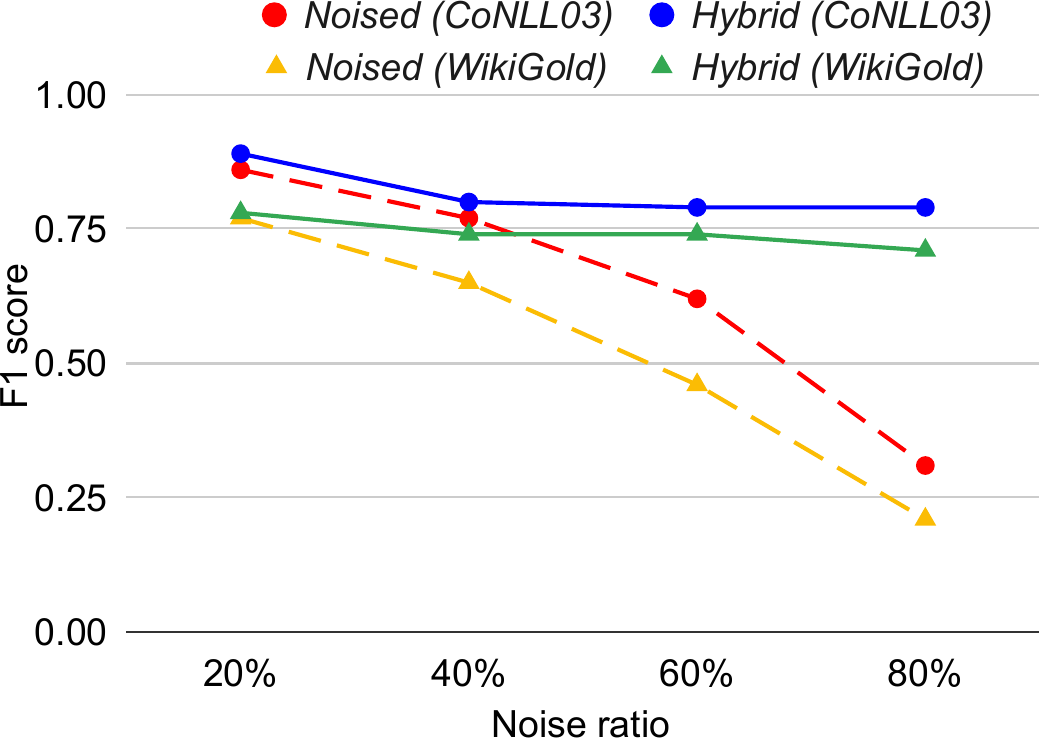}
    \caption{Comparison of model performance when using noised and datasets annotated by Hybrid approach.}
    \label{fig:hybrid}
    \vspace{-2ex}
\end{figure}

\subsection{Label Mixing in LLM-Based Annotation}
Addressing the issue, as mentioned earlier, of multiple labels being assigned to a single word, we extended the Mixup~\citep{zhang2018mixup} technique, traditionally used in image processing, to NER tasks.
Mixup is a data augmentation method that improves model performance and robustness by mixing two inputs and outputs.
Our approach involves blending two different labels for a single token to create new labels. This method is formally represented as follows:

\vspace{-3ex}
\begin{equation*}
\boldsymbol{y}_{\text{mix}} = \lambda \times \boldsymbol{y}_i + (1 - \lambda) \times \boldsymbol{y}_j
\end{equation*}

% https://arxiv.org/pdf/1710.09412.pdf
Where $\boldsymbol{y}_i$ and $\boldsymbol{y}_j$ are one-hot encodings of two different labels annotated entity types, and $\lambda \in [0,1]$ is a parameter determining the mixing ratio. Following the mixup technique~\citep{zhang2018mixup}, $\lambda$ is drawn from the beta distribution ($\alpha=\beta=0.2$).
Compared to randomly selecting one label from multiple labels assigned to a single word, this technique contributes to a more nuanced understanding of tokens that could belong to multiple entity types, enhancing the dataset's richness and classification flexibility.

\section{Experiments}

% \usepackage{multirow}
% \renewcommand{\arraystretch}{0.9} % 行間隔を縮小
\renewcommand{\arraystretch}{1.2}
\begin{table}[tb]
  \small
  \centering
  \begin{tabular}{crrr}
    \toprule \textbf{Budget}        & \textbf{\#Manual} & \textbf{\#LLM} & \textbf{F1 Score} \\
    \midrule \multirow{4}{*}{\$38}  & 87                & 0              & 0.02              \\
                                    & 70                & 702            & 0.81              \\
                                    & 35                & 2,106          & 0.84              \\
                                    & 0                 & 3,510          & {\bf 0.85}        \\
    \midrule \multirow{4}{*}{\$152} & 351               & 0              & 0.63              \\
                                    & 280               & 2,808          & 0.85              \\
                                    & 140               & 8,424          & {\bf 0.87}        \\
                                    & 0                 & 14,041         & 0.86              \\
    \midrule \multirow{3}{*}{\$608} & 1,404             & 0              & 0.86              \\
                                    & 1,263             & 5,616          & {\bf 0.87}        \\
                                    & 1,123             & 11,232         & {\bf 0.87}        \\
    \bottomrule
  \end{tabular}
  \vspace{2mm}
  \caption{Comparison between different data count balances from manual and LLM-based
  data under the fixed budgets.}
  \vspace{-3ex}
  \label{tab:budget_comparison}
\end{table}

\textbf{Setup.}
We conducted experiments using two distinct datasets. 
Firstly, we employed the CoNLL03 \citep{tjong-kim-sang-de-meulder-2003-introduction}, recognized as a standard dataset for NER. 
Additionally, acknowledging recent advancements in this field that have led to a saturation in performance improvements on the CoNLL03, we also included the WikiGold~\citep{balasuriya-etal-2009-named}. The WikiGold dataset was specifically chosen due to its higher level of challenge and the observation that performance levels achieved on it have not yet paralleled those on the CoNLL03. This was split into training, validation, and test sets in a 7:1:2 ratio for the experiments.

For the automated annotation process, we leveraged ChatGPT (\texttt{gpt-3.5-turbo-0613})\footnote{\href{https://chat.openai.com}{https://chat.openai.com}}, and we set the number of few-shot examples at 32.

Finally, to validate the effectiveness of our annotation approach, we fine-tuned a sequence classification model based on a pre-trained BERT~\citep{devlin-etal-2019-bert}, utilizing the datasets annotated through this methodology. The evaluation of the NER task is in line with \citet{tjong-kim-sang-de-meulder-2003-introduction}, and we present weighted average F1 scores in this paper.

A detailed description of the hardware and software configurations is also provided in \Cref{sec:appendix-experiment-setup} to ensure reproducibility.

% \subsection{Results}

\textbf{Noise Recovery with LLM-based Annotations.}
We assessed the capability of LLM-based annotations to recover from noisy data. \citet{han2023information} pointed out some annotation errors, such as missing annotations or label switching, in the CoNLL03 dataset~\citep{tjong-kim-sang-de-meulder-2003-introduction}. We prepared noised data that reproduces this problem by removing annotations for some entity. We show that it can be improved by combining LLM-based annotation in the previously described manner. We experimented with 20\%, 40\%, 60\%, 80\% entity annotations removed. By comparing the noised datasets to those enhanced through our proposed methodology, we established a baseline for improvement. Figure \ref{fig:hybrid} presents a comparative analysis of these datasets, highlighting the efficiency of our approach to noise reduction. It was shown that even when noise is severe, it can be significantly improved by combining LLM-based annotations. 

\textbf{Comparison Under the Same Budget.}
We evaluated our approach under identical budget constraints by drawing parallels with the study of~\citet{ding-etal-2023-gpt}. Our experiments were conducted on three different budgets. We created composite datasets including different numbers of samples from manual and LLM-based annotations. The middle budget (\$152) is the cost of LLM-based annotation of all training data of the CoNLL03 dataset, and the costs of a quarter and a quadruple of it are assumed for small (\$38) and large (\$608) budgets, respectively. The relationship between the balance of the number of data in each budget and performance is shown in Table \ref{tab:budget_comparison}.

When the budget is small (\$38), training is not possible with only manual data due to the small amount of data. Table \ref{tab:budget_comparison} shows that it is thought that performance can be improved by using LLM-based annotations to increase the training data. For the middle budget (\$152), the model's performance was reduced using only the manual data; using just 2,808 LLM-based annotations showed a significant improvement. In addition, the performance of the Hybrid annotations was higher than that of LLM-based alone, suggesting that the use of both annotations contributes to the performance. For the large budget (\$608), the performance is sufficient with 1,404 manual data, but a slight performance improvement was confirmed by reducing the manual data and using LLM-based as well. The results show that the composite datasets can increase the number of data at a low cost and improve the performance of downstream tasks under all budgets.

\begin{figure}[t]
  % \vspace{-3mm}
  \centering
  \includegraphics[width=1.0\linewidth]{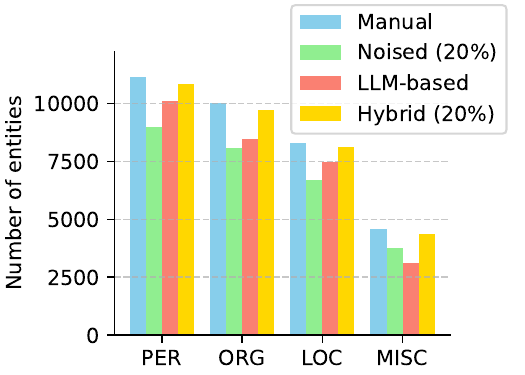}
  \caption{Number of annotations for each entity type in the manual, noised, and LLM-based datasets.}
  \vspace{-2ex}
  \label{fig:entity_comparison}
  % \vspace{-3mm}
\end{figure}

\textbf{Addressing the Imbalanced Class Issue.}
\Cref{fig:entity_comparison} shows that there is a difference in the number of annotations for each entity type. This imbalance issue is even greater with annotation by LLM.
Also, the performance difference of LLM-based annotation per entity type is large compared to manual annotation.
In terms of the performance difference between labels, the variance of F1-scores when using LLM-based annotation is larger than that of manual annotation. 
% To tackle the class imbalance issue, we motivated the use of our proposal, label mixing. We present the comparison between LLM-based annotation and label mixing in \Cref{fig:label_mixing}.
Label mixing reduced the performance difference between entity types. Therefore, we believe that soft labels, according to possible entities, will prevent model training in a biased manner, unlike one-hot labels.

% \begin{figure}[tb]
%     \centering
%     \begin{minipage}[b]{0.5\linewidth}
%         \centering
%         \includegraphics[width=\linewidth]{figs/conll03_labelmixing-crop.pdf}
%     \end{minipage}%
%     \begin{minipage}[b]{0.5\linewidth}
%         \centering
%         \includegraphics[width=\linewidth]{figs/wikigold_labelmixing-crop.pdf}
%     \end{minipage}
%     \caption{Improvement of each entity type and micro average with label mixing.}
%     \label{fig:label_mixing}
% \end{figure}

\renewcommand{\arraystretch}{1.2}
\begin{table}[tb]
  \small
  \centering
  \resizebox{\columnwidth}{!}{%
    \begin{tabular}{l|rr|rr}
      \toprule
      & \multicolumn{2}{c|}{CoNLL03} & \multicolumn{2}{c}{WikiGold} \\
      Annotations & mean                         & variance                    & mean & variance \\
      \midrule \textbf{\begin{tabular}[c]{@{}l@{}}LLM-based\\ w/o label mixing\end{tabular}} & 0.83                         & 0.011                       & \textbf{0.63} & 0.043    \\
      \textbf{\begin{tabular}[c]{@{}l@{}}LLM-based\\ w/ label mixing\end{tabular}}           & \textbf{0.84}                         & \textbf{0.008}                       & \textbf{0.63} & \textbf{0.032}    \\ \midrule
      Manual                                                               & 0.89          & 0.005          & 0.80          & 0.009          \\
      \bottomrule
    \end{tabular}
  }
  \caption{Comparison on the average F1-score and variance between before and after label mixing.}
  \vspace{-2ex}
  \label{tab:label_mixing_variance}
\end{table}

\renewcommand{\arraystretch}{1.2}
\begin{table}[tb]
  \small
  \centering
  \resizebox{\columnwidth}{!}{%
    \begin{tabular}{l|ccc|ccc}
      \toprule     & \multicolumn{3}{c|}{CoNLL03} & \multicolumn{3}{c}{WikiGold} \\ %\cline{2-7}
                   & \multicolumn{1}{l}{ORG}      & \multicolumn{1}{l}{LOC}     & \multicolumn{1}{l|}{MISC} & \multicolumn{1}{l}{ORG} & \multicolumn{1}{l}{LOC} & \multicolumn{1}{l}{MISC} \\
      \midrule PER & 72                           & 37                          & 15                        & 13                      & 5                       & 1                        \\
      ORG          & -                            & 180                         & 59                        & -                       & 44                      & 41                       \\
      LOC          & -                            & -                           & 158                       & -                       & -                       & 5                        \\
      \bottomrule
    \end{tabular}%
  }
  \caption{The number of entities overlapping among classes in the CoNLL03 and WikiGold dataset.}
  \vspace{-4ex}
  \label{tab:number_of_overlapping_entities}
\end{table}

% In addition to the analysis above, we further explore the impact of LLM-based annotations and label mixing on the CoNLL03 and WikiGold datasets, particularly focusing on the number of entities that overlap between two classes. 
\Cref{tab:number_of_overlapping_entities} presents a detailed quantification of these overlapping entities between different entity types when annotated by LLM. 
% The fact that the entities overlapping between classes constitute only 521 cases (2.65\%) of the CoNLL03 dataset and 109 cases (6.80\%) of the WikiGold dataset, and yet the application of label mixing to these specific data points results in a performance improvement across the dataset, is a suggestive observation. This observation implies that the strategic application of label mixing to even a small subset of data can improve the performance balance between entity types. 
Moreover, the fact that the number of entities overlapping between classes such as LOC and MISC with ORG is comparatively higher than other combinations, while there is less overlap between PER and MISC, indicates a potential challenge in the discriminatory capability of LLM-based annotations for the former combinations. These observations support the effectiveness of label mixing as a technique to buffer these shortcomings, suggesting that it serves as a valuable strategy for improving the robustness and performance of NER models.
\section{Discussion and Conclusion}

Our study extends the use of LLMs toward the annotation process. 
We addressed NER annotation challenges of missing labels and proposed novel solutions. 
Our integration of LLM-based annotation with manual annotation significantly enhances performance of NER task, especially when using noised datasets or in limited budgets.
We have shown that our proposed label mixing address the performance degradation due to biases and imbalances.
% Our approa leverages LLMs' evolving strengths~\citep{zhao2023survey,min2023recent}, continuing to improve in the future.
Our approach capitalizes on the capabilities of LLMs, which are expected to continue advancing in the future~\citep{zhao2023survey,min2023recent}.
% and introduces label mixing to address biases and imbalances from LLM-based annotations. 
% Our work illuminates LLMs' role in enhancing NER, underscoring the need for ongoing adaptation and improvement in annotation methodologies.

\section{Ethical Considerations}

In our research, we focus not on generating text, but on the automation of annotations. This approach places concerns about creating potentially harmful sentences outside our scope of study. However, while our primary objective is to annotate entities with labels, it is essential to consider the implications of certain annotations, such as the toxicity of entities, especially when these annotations pertain to specific demographic groups.

This risk stems from biases inherent in the LLMs used for annotation, necessitating continuous monitoring.

\section{Limitations}

Finally, we would like to mention the limitations of our work. 
As the first limitation, we have only validated our proposed methods with relatively small datasets, CoNLL03 and Wikigold, thus requiring validation with larger NER datasets, such as OntoNotes 5.0 \citep{AB2/MKJJ2R_2013}.

Another limitation is that when using the ChatGPT (gpt-35-turbo-0613) API, some regions and organizations are not generated due to their alignment. It is conceivable that this kind of alignment-based distribution bias could affect NER task performance.

The final limitation involves the LLM-based annotation method for NER tasks, which requires computational costs proportional to the number of entity types. Thus, algorithms independent of entity type count are needed.

% \section*{Acknowledgments}

\bibliography{main}
% \bibliography{custom}

\newpage
\clearpage
\appendix

\section*{Appendix}
\section{Details of Experiment Setup}
\label{sec:appendix-experiment-setup}
% \begin{description}[leftmargin=*,itemsep=0pt,topsep=0pt]

% \item[Hardware Configurations:]
\subsection{Hardware Configurations:}
In our experiments, we used virtual machines on Google Cloud. The detailed machine specifications are as follows:
\begin{itemize}
\setlength{\itemsep}{0pt}
\setlength{\parskip}{0pt}
    \item Machine type: \texttt{n1 standard 4}
    \item Memory: 15 GB
    \item virtual CPU: \texttt{Intel Broadwell} x 4
    \item GPU: \texttt{NVIDIA T4} x 1
\end{itemize}

\vspace{3mm}
% \item[BERT Configurations:]
\subsection{BERT Configurations:}
The configurations of BERT used for the performance evaluation are as follows:
\begin{itemize}
\setlength{\itemsep}{0pt}
\setlength{\parskip}{0pt}
  \item Pre-trained model: \texttt{bert-model-unbased} \footnote{\href{https://huggingface.co/bert-base-uncased}{https://huggingface.co/bert-base-uncased}}
  \item Learning rate: $1e-05$
  \item Upper limit for norm of gradients: $10$
  \item Batch size: $64$ for training, $32$ for inference 
  \item Max length of token: $128$
\end{itemize}

% 学習は、3エポック続けて Accuracy が変化しなくなったところで終了することとした。
Each training was terminated when the F1 score on the validation set remained unchanged for three consecutive epochs.

% NER では各単語単位にラベルを割り振る必要があるが、BERT のトークナイザーでは一つの単語を複数の単語に分割する場合がある。
% そこで本研究では、単語が複数のトークンに分割された場合、先頭のトークンに対するラベルを単語全体のトークンとして最小している。
In NER, labels must be assigned at the word level, but BERT's tokenizer often splits a single word into multiple tokens.
Therefore, in this study, when a word is split into multiple tokens, when a word is split into multiple tokens,
the model adopts the label for the first token as the label for the entire word.

\subsection{Code Implementation}
The experimental code for this study was implemented using Copilot.\footnote{\url{https://github.com/features/copilot}}

% \end{description}

\section{Datasets}
% \begin{description}[leftmargin=*,itemsep=0pt,topsep=0pt]

% \item[CoNLL03~\citep{tjong-kim-sang-de-meulder-2003-introduction}:]
\subsection{CoNLL03:}
The CoNLL-2003 dataset~\citep{tjong-kim-sang-de-meulder-2003-introduction} is designed to recognize four types of named entities, including person names, place names, organization names, and other miscellaneous entities. The data file consists of four columns: word, part-of-speech tag, syntactic chunk tag, and named entity tag, with each item separated by a space. The named entity and syntactic chunk tags use the I-TYPE format \footnote{This tag means "Inside" and is assigned to words that are within an entity of a specific type (TYPE). That is, this tag is used when a word is part of a specific entity and indicates the type of that entity. For example, "I-PER" indicates that the word is part of a person's name (Person).}, and the B-TYPE \footnote{The B in B-TYPE means "Beginning" and is the tag given to the first word of the second entity when directly followed by an entity of the same type. This tag is used to indicate that a new entity is beginning and to distinguish it from the previous entity. For example, "B-PER" means that the new person's name begins with this word.} tag is introduced to indicate the start of a new phrase.

\vspace{3mm}
% \item[WikiGold~\citep{balasuriya-etal-2009-named}:]
\subsection{WikiGold:}
The Wikigold dataset \cite{balasuriya-etal-2009-named} is a manually annotated corpus for named entity recognition consisting of a small sample of Wikipedia articles. Words are labeled for each of the four named entity classes (LOC, MISC, ORG, and PER) of CONLL-03 described above.

% \end{description}

% \begin{table*}[t]
%     \centering
%     \begin{tabular}{l|ccccc}
%     \hline
%     CoNLL03 & PER & ORG & LOC & MISC & Micro Ave \\ \hline
%     % Manual & 0.97 & 0.88 & 0.93 & 0.78 & 0.91 \\
%     LLM-based & 0.93 & 0.84 & 0.90 & 0.66 & 0.86 \\
%     LLM-based (label mixing) & 0.95 & 0.82 & 0.89 & 0.71 & 0.87 \\ \hline
%     WikiGold & PER & ORG & LOC & MISC & Micro Ave \\ \hline
%     % Manual & 0.95 & 0.76 & 0.82 & 0.65 & 0.80 \\
%     LLM-based & 0.91 & 0.57 & 0.71 & 0.34 & 0.67 \\
%     LLM-based (label mixing) & 0.87 & 0.60 & 0.68 & 0.37 & 0.65 \\ \hline
%     \end{tabular}
%     \caption{Comparison between LLM-based annotation and the one using label mixing on CoNLL03 and WikiGold datasets.}
%     \label{tab:label_mixing}
%     \vspace{-3mm}
% \end{table*}

% \section{Related Work}

\section{Statistics of Overlapped Entities}

LLM-based annotations can overlap different classes on the same entity. We investigate the statistics of overlapped entities.

The fact that the entities overlapping between classes constitute only 521 cases (2.65\%) of the CoNLL03 dataset and 109 cases (6.80\%) of the WikiGold dataset, and yet the application of label mixing to these specific data points results in a performance improvement across the dataset, is a suggestive observation. This observation implies that the strategic application of label mixing to even a small subset of data can improve the performance balance between entity types.

\end{document}